\documentclass{article} 
\usepackage{iclr2016_conference,times}
\usepackage{url}
\usepackage{graphicx}
\usepackage{tabularx,ragged2e,booktabs,caption}
\usepackage[utf8]{inputenc}
\usepackage{epstopdf}

\title{8-Bit Approximations for Parallelism in Deep Learning}

\author{Tim Dettmers \\
The Faculty of Informatics\\
Universià della Svizzera italiana\\
Via Giuseppe Buffi 13, CH-6904 Lugano, Switzerland \\
\texttt{tim.dettmers@gmail.com} 
}

\begin{document}
 \iclrfinalcopy

\maketitle

\begin{abstract}
The creation of practical deep learning data-products often requires parallelization across processors and computers to make deep learning feasible on large data sets, but bottlenecks in communication bandwidth make it difficult to attain good speedups through parallelism. Here we develop and test 8-bit approximation algorithms which make better use of the available bandwidth by compressing 32-bit gradients and nonlinear activations to 8-bit approximations. We show that these approximations do not decrease predictive performance on MNIST, CIFAR10, and ImageNet for both model and data parallelism and provide a data transfer speedup of 2x relative to 32-bit parallelism. We build a predictive model for speedups based on our experimental data, verify its validity on known speedup data, and show that we can obtain a speedup of 50x and more on a system of 96 GPUs compared to a speedup of 23x for 32-bit. We compare our data types with other methods and show that 8-bit approximations achieve state-of-the-art speedups for model parallelism. Thus 8-bit approximation is an efficient method to parallelize convolutional networks on very large systems of GPUs.
\end{abstract}

\section{Introduction}
Deep learning is a field inherently driven by advances in computational processing \citep{schmidhuber2015deep}. Graphics processing units (GPUs) can accelerate deep learning by a factor of up to 10-20 compared to a CPU, and these speedups were integral in achieving breakthroughs in speech recognition and computer vision \citep{ciresan2012multi, dahl2012context,krizhevsky2012imagenet}. After these breakthroughs, GPUs found widespread use and many teams sought to accelerate the traininfple GPUs or computers \citep{chilimbi2014project,coates2013deep,dean2012large,wu2015deep}. To make deep learning applicable and scalable for large data sets, it is important to develop successful parallel deep learning algorithms.

The main difficulty in the parallelization of deep learning is the sequential nature of backpropagation, where the parameter updates must be fully completed before the next iteration of stochastic gradient descent can ensue \citep{rumelhart1988learning}. This creates an environment where transfer of parameters between GPUs and computers require high bandwidth and low latency for network communication. Network communication generally constitutes the major bottleneck in deep learning parallelism.

There are two major ways to increase the performance of a parallel deep learning algorithm: (1) Overlap communication and computation in such a way that most of the communication is done while waiting for a computation to finish; (2) Decrease the number or size of parameters needed to transfer;

In this paper we work on (2) and make the following contributions: 
\begin{itemize}
	\item We discuss all important hardware and software bottlenecks in model and data parallel deep learning algorithms on GPUs and GPU clusters
	\item We develop 8-bit gradient approximation data types and by applying them to MNIST, CIFAR10 and ImageNet show that the error rates remain unchanged despite this approximation
	\item We build a predictive model based on our experimental data and show that it predicts the speedup obtained by \citet{krizhevsky2014one} with a relative error of about 1\%. We use this model to show that 8-bit approximation leads to minor speedups in convolutional networks with 4 GPUs, but to a speedup of 50x and more for a system with 96 GPUs compared to up to 23x for 32-bit.
	\item We compare our algorithm with similar work and show that 8-bit approximation is able to circumnavigate problems with large batch sizes for GPU clusters and thus improves convergence rates in convolutional networks
	\item We show that 8-bit approximation sets the state-of-the-art for model parallelism in general
\end{itemize}
\section{Background}

 To understand the properties of a successful parallel deep learning algorithm, it is necessary to understand how the communication between GPUs works and what the bottlenecks for both model and data parallelized deep learning architectures are. First we look at the specific properties of data and model parallelism, and then we look at general bottlenecks in GPU-to-GPU communication.

\subsection{Data parallelism}

In data parallelism, the model is kept constant for all GPUs while each GPU is fed with a different mini-batch. After each pass the gradients are exchanged, i.e. synchronized with each GPU:
\begin{itemize} 	
	\item How it is done: For fully connected layers, data is split by the sample dimension, e.g. for 4 GPUs, a 1024x784 mini-batch is split into four 256x784 batches; for convolutional layers, the data is split by the sample dimension (better cross-validation error) or by the feature map dimension (decreases memory usage dramatically; slightly worse cross-validation error; makes architecture complicated)
	\item Infrequent synchronization: Parameters are synchronized (averaged) once after each full forward+backward pass
	\item Efficiency: Data parallelism is efficient when the model has few parameters, e.g. long short-term memory recurrent neural networks \citep{hochreiter1997long}; or the computational costs per parameter are high, e.g. convolutional layers in convolutional nets
	\item Scaling limitations: Current GPU implementations are optimized for larger matrices, hence data parallelism does not scale indefinitely due to slow matrix operations (especially matrix multiplication) for small mini-batch sizes ($< 128$ per GPU); convolution implementations that rely on matrix multiplication may suffer from this too; the larger the batch size the slower the convergence to a local minimum which is problematic for large systems
	\item Requires asymptotic accuracy: Good solutions can be found as long as the sequence of updates converges to the minimum asymptotically \citep{seide20141}
\end{itemize}

\subsection{Model parallelism}
In model parallelism, the data is kept constant for all GPUs while each GPU holds only a part of the full model:
\begin{itemize}
	\item How it is done: Distribute the parameters of a layer on multiple GPUs (split by input or output dimension); pass the same mini-batch through the distributed layer
	\item Frequent synchronization: Parameters are synchronized once for every layer; the outputs of the layer are either stacked or added together depending on the matrix operation
	\item Efficiency: Model parallelism is efficient when the layer has many parameters, e.g. in fully connected layers (because the parameter matrix is reduced by a factor equal to the number of GPUs)
	\item Scaling limitations: Poor performance for larger mini-batch sizes; the larger the mini-batch size, the larger the matrix that needs to be synchronized across GPUs
	\item Requires numerical accuracy: Outputs must be precise as small deviation may lead to large errors in later layers; this is similar to the exploding gradient problem \citep{hochreiter2001gradient}
\end{itemize}

\subsection{General bottlenecks}

\subsubsection{PCIe switches}

PCI Express (PCIe) is built like an ordinary network, where pairs of two PCIe slots share a common switch which can serve one outgoing and incoming connection simultaneously. Thus only a single device in a device-pair can communicate with another device-pair at any given time. This holds for both GPUs and InfiniBand cards. Thus PCIe switches need to be taken into account to obtain optimal performance in a multi-GPU system.

\subsubsection{Bandwidth}
GPUs within a computer communicate by using the PCIe interface which offers practically about 14 GB/s bandwidth when it contains 2 GPUs and about 7 GB/s bandwidth when a computer contains more than 2 GPUs. 

GPUs between computers usually communicate by using InfiniBand network cards which have a practical bandwidth of 3-7GB/s (quad-data-rate (QDR) and fourteen-data-rate cards (FDR), respectively).

Communication is the main bottleneck in deep learning which can be illustrated with a simple example: AlexNet is a convolutional network with about 60 million parameters \citep{krizhevsky2012imagenet}; a full forward-backward pass is completed in under 100ms for current generation GPUs\footnote{https://github.com/soumith/convnet-benchmarks}. When implementing naive data parallelism with 4 GPUs, we need to synchronize 60 million parameters with the 3 other GPUs. Due to PCIe switches over which GPU pairs have to send their messages, this takes as long as sending 4 messages between 2 unpaired GPUs. Now 60 million 32-bit parameters take up 0.223GB of memory, which at 7GB/s take 32ms for one transfer or 128ms for a full synchronization which means that naive data parallelism on four GPUs would be {\it slower} than running the network on a single GPU. Thus using more GPUs or computers is not always beneficial. This also demonstrates the need for high communication bandwidth in large-scale deep learning.

\subsubsection{Latency}

PCIe latency between messages is usually in the microsecond regime and does not increase significantly with message size. Thus PCIe latency is negligible. However for larger systems with multiple computers that make use of InfiniBand the latency can be a considerable bottleneck, especially for collective communication (one-to-all, all-to-one, all-to-all) on clusters \citep{sur2005high, singh2012optimizing}. Latency for current InfiniBand systems (FDR) increases exponentially, and above 512 kilobytes of data it becomes unmanageable for GPU clusters with more than a dozen of nodes (latency $>$ 0.5ms per message). Thus it is imperative to keep messages relatively small or otherwise performance is crippled considerably. Latency is the biggest bottleneck in deep learning systems for large-scale GPU clusters.

\subsection{Optimal parallelism for convolutional nets}

The currently best known method to parallelize convolutional nets on any number of $K$ GPUs or computers is to use data parallelism in the convolutional layers and model parallelism in the fully connected layers \citep{krizhevsky2014one}. However, we only perform a single model parallel step, but $K$ such steps with a $K$th of the full batch size. So after the convolutional layers -- that is before the fully connected layers -- 1/$K$th of each batch (from hereon sub-batch) is distributed across all GPUs and then a model-parallel forward and backwards pass up to the convolutional layers is performed. During each partial forward-backward pass, the next sub-batches are distributed across all GPUs, thus hiding the communication time for all incoming sub-batches under the forward-pass computation time. The same procedure can be applied to synchronize the fully connected activities during model parallelism, which thus hides almost all necessary communication. 

During this procedure, the gradients for the first partial forward-backward passes may be used to update the fully connected layers directly rather than waiting for the other $K$ mini-batches for an average gradient. These multiple updates improve the time to convergence and can be used to hide further communication.

This whole process is repeated $K$ times until all $K$ sub-batches have completed a full pass up to the convolutional layers. After this model parallelism step in the fully connected layers, normal data parallelism ensues.

\section{8-bit approximation}

We seek to develop an approximation for gradients which is small yet has sufficient accuracy to be usable for both data and model parallelism. The standard size of gradients in deep learning is currently 32-bit, which is the smallest practical dimension for floating point numbers on GPUs, as CUDA only supports 32 and 64-bit floating point arithmetic. We chose to use 8-bit for our gradient approximation data type because (1) it is easy to handle as we can store it in 8-bit unsigned chars, and (2) we reasoned that less than 8 bits would have insufficient accuracy for model parallelism, since existing literature suggested that less than 8 bits can induce considerable reduction in accuracy \citep{courbariaux2014low}. 

\subsection{Designing 8-bit data types}

From these 8 bits, one bit is reserved for the sign of the number, while the rest can be used for exponent and mantissa. 

The problem with the mantissa is that our range of precision is very limited. With a 3-bit exponent the mantissa will hold values between 0 and 15 and as such decimal values over 15 will have a poor approximation and thus a large error. For example the numbers ending in 2 to 2.499 will be approximated by numbers ending in $2$, yielding an average relative error of 22.5\% in this range.

In order to decrease this error, we can use the bits of the mantissa to represent a binary tree with interval $(0.1,1)$ which is bisected according to the route taken through the tree; the children thus represent the start and end points for intervals in a bisection method. With this method we can cover a broader range of numbers with the mantissa and can thus reduce the average relative error. \\\\
We can decrease this error further with another method: We can use additional bits from the exponent and introduce a dynamic exponent instead. This dynamic exponent may use between 0 to 7 bits, where the number $n$ of leading 0 bits represents the exponent $10^{-n}$; the first bit which is set to 1 is a flag which indicates that the next bits are part of the binary bisection tree. With this format we lose the ability to represent large exponents (a maximum of $10^{-6}$ instead of $10^{-7}$) and we lose one bit for the binary bisection tree, but we gain the ability to approximate numbers with large absolute value with smaller error (e.g. 0.2345678 approximated as 0.236719 instead of 0.23125 or 0.2, respectively), while retaining the ability to approximate numbers with small absolute value that have few significant digits (0.0000234 approximated as 0.000019). However, one downside is that our approximation of values below $10^{-3}$ loses some accuracy because the zeros (e.g. 2 zeros and 1 flag bit) contain less information than the equivalent bits. However, gradients and activations with absolute size greater $10^{-3}$ are arguably more important for learning than gradients and activations below $10^{-3}$ because they simply have a larger effect. Thus this data type should yield better training and predictive performance for deep learning algorithms. 

To increase the accuracy for data and model parallelism respectively, we can introduce fitting offsets for the exponent. Because model parallelism often induces larger activation function values which have high variance (especially for piecewise-linear functions), an exponent offset of $10^2$ to $10^4$ is desirable (exact value depends mainly on the used nonlinear activation function). 

For the data type with dynamic exponent, we can instead normalize a matrix by dividing by its absolute maximum value and thus transform all numbers to be in the range $[1,0]$ which is then suitable for the bisection method; upon decompression to a 32-bit float we just multiply each approximated value by the absolute maximum value to renormalize it. Using this method with a 7-bit bisection tree with the range [0,1], we receive a data type which is equivalent to linear quantization \citep{vanhoucke2011improving}.

\begin{figure}[h]
	\begin{center}
		\fbox{\includegraphics[width=1\linewidth]{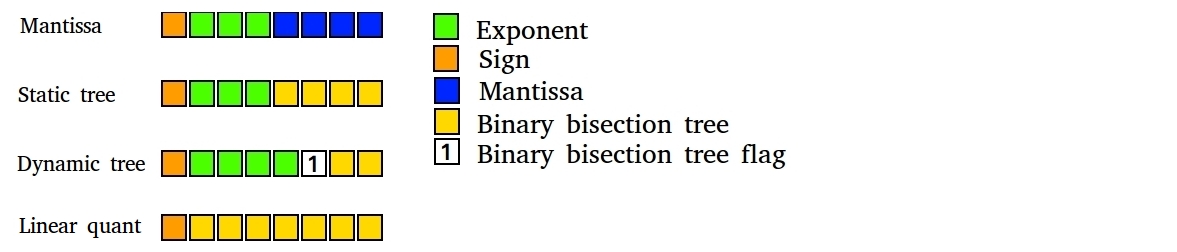}}
	\end{center}
	\caption{Anatomy of the four different 8-bit data types. Note that the dynamic data type shown here is a specific example and the number of bits for the exponent and for the tree varies between individual floating point numbers.}
\end{figure}

\subsection{Implementation and computational performance}

The fastest implementation for 8-bit compression and decompression we could think of is to use a binary search on a sorted table of all positive 128 values in shared GPU memory and keep track of the sign of the respective number. Shared memory is about 100 times faster than global GPU memory and thus a binary search in shared memory is very fast. 

In our implementation we used one thread per number in a binary search. Additional parallelization is easily possible by dividing the table into $n$ intervals, where $n$ is the number of threads per number. However, the necessary thread synchronization is expensive and performance gains are probably negligible compared to the additional resource costs (threads). 

For decompression, the 32-bit values are read into shared memory and we lookup the 32-bit value for the respective 8-bit value. Here we use one thread per number/lookup.

On average, these algorithms perform compression and decompression in 1 and 0.5 nanoseconds per number, respectively, as measured on a NVIDIA GTX Titan.

Implementations of our 8-bit approximation algorithms are available online \footnote{https://github.com/TimDettmers/clusterNet/; contact me if you need help with integrating the functions into your library}.

\subsection{Theoretical speedup}
\begin{table}[h]
	\caption{Predicted speedups for AlexNet for 4 and 96 GPUs.}
	\label{sample-table}
	\begin{minipage}
		{\linewidth}
		\centering
		\begin{tabular}{ cccc}
			\toprule[1.5pt]
			GPUs &  Sub-batch size &   \multicolumn{2}{c}{Speedup}    \\		 
			& &    32-bit & 8-bit		 \\
			\midrule
			4 & 128 & 3.53x & 3.67x \\
			96 & 128 & 13.4x & 13.8x		\\
			96 & 256 & 20x & 24.4x \\	
			96 & 512  & 23x & 39.3x \\		
			96 & 1024 & 14.8x & 48.6x \\			
			96 & 2056 & 11.3x & 50.6x \\	
			96 & 12288 & 1.3x & 9.7x	\\			
			\bottomrule[1.25pt]
		\end{tabular}
		\par
		\bigskip
	\end{minipage}
\end{table}

We measured the average total transfer time (compression, transfer, and decompression) for our techniques and compared them to 32-bit transfers between GPUs. We measured this time on a board equipped with 4 GPUs which yields 8 PCIe 3.0 lanes for each GPU and thus a theoretical bandwidth of about 8GB/s; however, bandwidth for small messages is usually considerably lower. The algorithms were run on two NVIDIA GTX Titans. Each matrix was transfered 100 times and the average total transfer time was measured. We used the message passing interface (MPI) implementation provided by OpenMPI 1.8.5, which uses low level CUDA routines to enable GPU-to-GPU communication without the help of the CPU. MPI is commonly used to parallelize algorithms on GPU clusters.

We then used these measurements to build a predictive model for speedup for different hardware configurations. We validated our model by predicting the speedup for \citet{krizhevsky2014one} which we could predict with a relative error of about 1\%. We also created a theoretical model for GPU clusters based on publicly available benchmark data on InfiniBand systems. The predictive model was then used to generate the theoretical speedups in Table 1. For more information on the predictive model see the Appendix where the model is derived.

From Table 1 we can see that the sub-batch scheme is essential to obtaining good speedups since we get almost no speedups if we use the full batch size of 12288. We also see that 8-bit approximation increases the speedup considerably as the sub-batch size gets larger. While speedups are larger for larger sub-batches, we will have slower convergence as shown by \citet{krizhevsky2014one}, which is similar to how a large batch size for any model increases the time to convergence. So this scheme yields good scaling, yet overcomes problems with large batch sizes, as is typical for full data parallelism approaches such as 1-bit quantization which is discussed further in section 4.2.

Furthermore, since no other approximation method is known to us which is usable with model parallelism, we conclude that 8-bit approximation is the currently best method to speed up communication for model parallelism. 

\subsection{Approximation error}

We tested the approximation error of our data types on multiple distributions and on the gradients (data parallelism) and activations (model parallelism) on MNIST (see Table 2). We calculated the mean absolute and relative error from a sample of size 25 million numbers drawn from normal distributions $N(\mbox{mean},\mbox{variance})$, and the uniform distribution $U(0,1)$. The approximation of $N(0,10^2)$ was done by using an exponent offset of $10^2$ while other numbers used an exponent offset of $10^1$. For the dynamic tree and linear quantization the sample was normalized by division by the maximum absolute value and then denormalized after compression. \\\\
As one can see from Table 2, the 8-bit dynamic tree provides the overall best performance for approximation numbers for random distributions and for parallelism on MNIST. 

\begin{table}[h]
	\caption{Approximation errors on different data sets, and for different layers on MNIST.}
	\label{sample-table}
	\begin{minipage}
		{\linewidth}
		\centering
		\begin{tabular}{ cccc}
			\toprule[1.5pt]
			Distribution & Data type & Mean absolute error & Mean relative error in \%\\\midrule
			$U(0,1)$ & 8-bit dynamic tree & {\bf0.00004} & {\bf 1.39}  \\
			- & 8-bit linear quant & 0.0024 & 2.16\\
			- & 8-bit mantissa & 0.0017 & 7.22\\
			- & 8-bit static tree & 0.0007 & 4.82\\\midrule	
			$N(0,1)$ & 8-bit dynamic tree & 0.0005 & {\bf2.46}\\
			- & 8-bit linear quant &  {\bf0.0004} & 6.47 \\
			- & 8-bit mantissa & 0.0104 & 6.89 \\
			- & 8-bit static tree & 0.0093 & 6.3 \\\midrule
			$N(0,10^2)$ & 8-bit dynamic tree & 0.049 & {\bf2.49} \\
			- & 8-bit linear quant & {\bf0.041 }& 6.44 \\
			- & 8-bit mantissa & 1.066 & 7.19\\
			- & 8-bit static tree & 0.926 & 6.28 \\\midrule
			$N(0,0.2^2)$ & 8-bit dynamic tree & 0.000018 & {\bf2.45} \\
			- & 8-bit linear quant &{\bf 0.000015} & 6.15 \\
			- & 8-bit mantissa & 0.00055 & 8.26\\
			- & 8-bit static tree & 0.00022 & 6.58 \\\midrule
			MNIST model parallel & 8-bit dynamic tree & {\bf 0.025$\>$/$\>$2$\>$/$\>$3321 }& {\bf0.5/2/2} \\
			- & 8-bit linear quant & 0.02$\>$/$\>$2.6$\>$/$\>$3387 & 1/2/3 \\
			- & 8-bit mantissa & 1$\>$/$\>$70$\>$/$\>$265000  & 1.5/2/8\\
			- & 8-bit static tree & 0.45$\>$/$\>$50$\>$/$\>$187000 &  1/2/6 \\\midrule
			MNIST data parallel & 8-bit dynamic tree & {\bf 0.0005$\>$/$\>$0.004$\>$/$\>$0.95} &  {\bf3/4/5} \\
			- & 8-bit linear quant & 0.001$\>$/$\>$0.01$\>$/$\>$1.25 & 5/6/6 \\
			- & 8-bit mantissa & 0.02$\>$/$\>$0.1$\>$/$\>$55 & 4.5/4.5/6\\
			- & 8-bit static tree & 0.01$\>$/$\>$0.04$\>$/$\>$24&  4/4/5\\
			\bottomrule[1.25pt]
		\end{tabular}
		\par
		\bigskip
	\end{minipage}
\end{table}

For our tests on MNIST we used rectified linear units, a 784x1024x1024x10 architecture with dropout (0.2,0.3,0.3), a learning rate of 0.003 and RMSProp  \citep{tieleman2012lecture}. Usually we have 8-bit approximation for all incoming GPUs and 32-bit gradients for the local GPU. Since we only had one GPU available for the following experiments, we simulated training on a large GPU cluster by only using the pure 8-bit approximation gradient component by training on a single GPU -- so no 32-bit gradients or activations where used. \\
On MNIST, we found that the best test error of all four approximation techniques static tree, dynamic tree, linear quantization, and mantissa  did not differ significantly from the test error of 32-bit training for both data parallelism $F(4,4) = 0.71, p = 0.59$, and model parallelism $F(4,4) = 0.54, p = 0.71$ (F-test assumptions were satisfied); also the 99\% confidence intervals did overlap for all techniques. Experiments with logistic units revealed the same results. This indicates that on MNIST, 8-bit approximation does not degrade classification performance compared to 32-bit, and all 8-bit approximation techniques are similar in performance on MNIST for both model and data parallelism.

We also tested our data types on CIFAR10, where we used a convolutional network\footnote{http://code.google.com/p/cuda-convnet/wiki/Methodology} with two convolutional layers (64x5x5, 64x3x3) which were followed by max-pooling (3x3) and contrast normalization after each layer. These layers were followed by two locally connected convolutional layers (no weight sharing) and a final fully connected softmax layer.

We used both data parallelism (convolutional layers) and model parallelism (local layers) for this convolutional net and we found that test errors for all 8-bit data types and 32-bit training only differed by at most 2\% relative to each other, indicating that 8-bit approximation did not decrease performance.

We also applied the 8-bit dynamic tree data type to AlexNet on the ImageNet dataset. We used our approximation scheme for both model (fully connected layers) and data parallelism (convolution). 

\begin{figure}[h]
	\begin{center}
		\fbox{\includegraphics[width=0.7\linewidth]{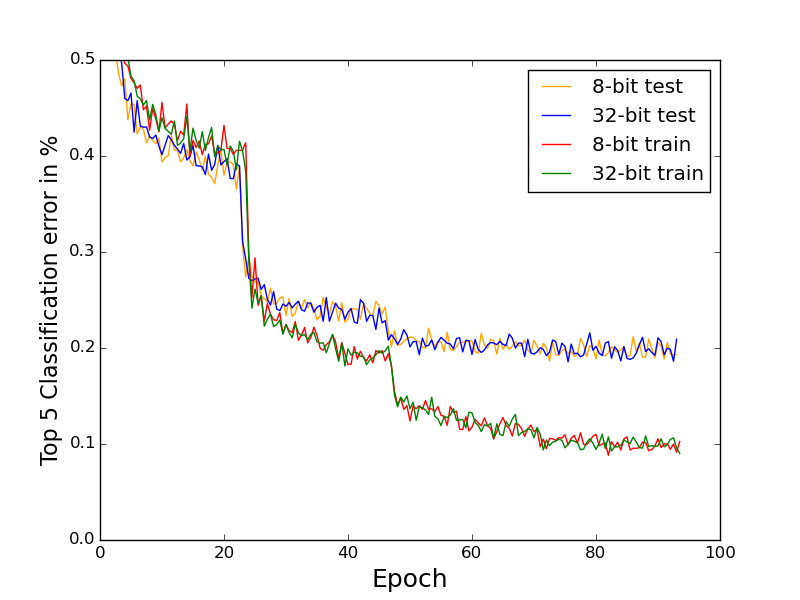}}
	\end{center}
	\caption{Classification train and test error for the 8-bit dynamic data type used in AlexNet on the ImageNet dataset.}
\end{figure}

Figure 2 shows that the 8-bit dynamic tree data type does not increase the misclassification error on the train or test set for convolutional nets. The final performance on the test set was comparable to the 32-bit model: 18.65\% and 18.55\% Top5-test-error for the 8-bit and 32-bit model, respectively.

\section{Comparison to other methods}

\subsection{Other sub-32-bit data types}

Dynamic fixed point data types are data types which use all their bits for the mantissa and have a dynamic exponent which is kept for collection of numbers (matrix, vector) and is adjusted during run-time. \citet{courbariaux2014low} used dynamic fixed point data types with 10-bit width for computation and 12-bit width for parameter updates to train a maxout convolutional network end-to-end. Their results on PI MNIST, MNIST, and CIFAR10 are about 20\% worse relative to the state of the art obtained by \citet{goodfellow2013maxout}.\\ 
In our work we show that we can use 8-bit gradients for the parameter updates without degrading performance. However, dynamic fixed point data types can also be used for end-to-end training and as such a combination of both methods might yield optimal performance. 

\citet{vanhoucke2011improving} used linear quantization, which is equivalent to our normalized data type with full binary tree and no exponent, in fixed point computation. They show that this data type can be used to achieve significant speedups on CPUs for a speech recognition task. Although our 8-bit data type with dynamic binary tree achieves better approximation, it cannot be used in fixed point computation and thus remains useful solely as an intermediate approximate representation.

\citet{gupta2015deep} used a 16-bit fixed point data type for end-to-end training of convolutional neural networks and showed that stochastic rounding improves the results significantly. We did not test stochastic rounding for our 8-bit data types, but it might also improve performance and thus allow the use of less than 8 bits for the approximation. This is left to explore for future research.

\subsection{1-bit quantization}
Another useful technique for data parallelism is 1-bit quantization which was introduced by \citet{seide20141}: In 1-bit quantization each 32-bit number of the gradient is quantized to a single bit by a quantization function. This quantization function maintains a cumulative quantization error which is used by the quantization function to smoothen out the error over time. The immediate error in quantization is too high to produce stable and accurate forward passes for model parallelism, but in data parallelism 1-bit quantization will converge to a local minimum seamlessly over time.

Compared to 8-bit approximation, 1-bit quantization performs well on medium sized systems that run fully connected architectures such as large fully connected networks for speech recognition \citep{strom2015scalable, seide20141}. For convolutional layers, 1-bit quantization has no advantage over 32 or 8 bits, as communication can be hidden under backward convolution operations even with 32-bit gradients and very large GPU clusters (see Appendix, section 5.2.2 for a worked example). 

Although no published example is known to us, 1-bit quantization should work flawlessly in the fully connected layers of convolutional networks. However, one problem with 1-bit quantization for large systems is the batch size which increases rapidly with the number of GPUs and slows down convergence \citep{seide20141,strom2015scalable, krizhevsky2014one}. To mitigate this problem \citet{seide20141} use adaptive batch size selection which determines the best batch size during runtime to improve convergence. Our 8-bit approximation scheme for convolutional networks does not suffer from this, as we use sub-batches in our model parallel pass, thus keeping the batch size small even for large GPU clusters (see Appendix, section 5.2).

Further improvement of 1-bit quantization comes from \citet{strom2015scalable} who only transfers floating point values of the gradient, which have an absolute value greater than a certain value $\tau$. Values that lie outside this interval are accumulated as residual gradients (similarly to the cumulative quantization error in 1-bit quantization) and applied over time. For sparse gradients, this procedure increases the compression factor considerably and thus decreases the time needed for communication. The problem of slow convergence with very large batch sizes should remain for very large systems, as observed by \citet{krizhevsky2014one}. However \citet{strom2015scalable} observed {\it improvement} in convergence rates with larger batch sizes but does not offer an explanation for this effect. 

\section*{Conclusion}

Here we have shown that approximation of 32-bit floating point numbers with 8 bits can speed up communication in parallel training of deep learning architectures considerably for large GPU clusters while retaining predictive performance. We have shown that the dynamic tree data type is able to approximate random numbers better than other known data types, but that during training all approximation techniques seem to perform equally well. We also showed that model parallelism in convolutional networks in conjunction with sub-batches works very well even for large GPU clusters and avoids the problem of large batch sizes that constitute problems for techniques like 1-bit quantization. Since no other approximation technique has been demonstrated for the compression of model parallel activations we set the state-of-the-art for model parallelism.

We expect that further important advances in parallel computing for deep learning will come from new hardware (3D GPU memory, EDR InfiniBand adoption, hardware based InfiniBand multicast) and new algorithms which maintain the performance of backpropagation while providing qualities which make them easier to parallelize. 

\bibliography{iclr2016_conference}
\bibliographystyle{iclr2016_conference}

\newpage
\section*{Appendix}

\section{Theoretical models for parallelism}
\subsection{A theoretical single node model for 4 GPUs}

\subsubsection{Benchmarking an architecture for single node 4 GPU parallelism}

We used NervanaGPU\footnote{https://github.com/NervanaSystems/nervanagpu} benchmarks to generate baseline and parallel time data for the AlexNet architecture as implemented in \citet{krizhevsky2014one}. NervanaGPU requires a Maxwell GPU or newer, but other deep learning libraries\footnote{https://github.com/soumith/convnet-benchmarks} exist with which benchmarking can also be easily handled. We benchmarked the timing of convolutional kernels, pooling operations, and matrix multiplications in fully connected layers for both model parallelism (output size divided by 4) and no parallelism. See Table 3 and Table 4 for the generated benchmark data. Note that the timings of the pooling operations are added to the convolutional timings; the pooling operations are in the order of 5-10\% of the timings of the respective preceding convolution. The transfer size is the size of the sub-gradient (one-fourth of the gradient, because we have 4 GPUs) or the size of the activation in model parallelism (batch size $\times$ output units / 4). The sync time is the time needed to transfer the buffer to all other GPUs (time equivalent to 4 messages needed due to PCIe switches). We benchmarked PCIe express transfer rates for different transfer buffer sizes and found that for most buffers we achieve a bandwidth of about 5GB/s (for smaller buffers this is less, for larger a bit more; the bandwidth saturates at 5.25GB/s for 4 GPUs on a Sandy Bridge Intel CPU). The total time is the time estimated by NervanaGPU benchmarks for the entire convolutional network and this number serves as the baseline from which we calculate speedup values. See my github repository\footnote{https://github.com/TimDettmers/public-data} for further data.

\begin{table}[h]
	\caption{Communication benchmark data for AlexNet. }
	\label{sample-table}
	\begin{minipage}
		{\linewidth}
		\centering
		\begin{tabular}{ cccc}
			\toprule[1.5pt]
			Layer/activation  & Transfer size (MB) &  \multicolumn{2}{c}{Sync time (ms)}  \\
			& & 32-bit & 8-bit \\
			\midrule
			conv11, stride 4, 3$\rightarrow$64 &   0.045 & 0.05 &-- \\
			conv5, 64$\rightarrow$192 &  0.29 &  0.45 & -- \\
			conv3, 192$\rightarrow$384 &  0.63 & 1  & --\\
			conv3, 384$\rightarrow$256 &   0.85 & 1.3 & -- \\
			conv3, 256$\rightarrow$256 &  0.56 & 0.9 & -- \\\midrule
			conv 3 activation &  1.125 & 0.9 & 0.4 \\
			fc 9216$\rightarrow$3072 &  1.5 & 1.1 &  0.5\\
			fc 3072$\rightarrow$3072 &  1.5 & 1.1 & 0.5\\
			fc 3072$\rightarrow$1000 &  0.5 & 0.4& 0.25\\\midrule
			
			\bottomrule[1.25pt]
		\end{tabular}
		\par
		\bigskip
	\end{minipage}
\end{table}

\begin{table}[h]
	\caption{Computation benchmark data for AlexNet. }
	\label{sample-table}
	\begin{minipage}
		{\linewidth}
		\centering
		\begin{tabular}{ ccc}
			\toprule[1.5pt]
			Layer/activation & Baseline time (ms) & Parallel time (ms)   \\
			& Fprop / Bprop / Update &  Fprop / Bprop / Update \\
			\midrule
			conv11, stride 4, 3$\rightarrow$64 & 4 / 1 / 3.5 & --   \\
			conv5, 64$\rightarrow$192 & 10 / 11 / 11 &  --  \\
			conv3, 192$\rightarrow$384 & 5 / 6 / 5 & -- \\
			conv3, 384$\rightarrow$256 & 6.5 / 5.5 / 6.5 &  -- \\
			conv3, 256$\rightarrow$256 & 4 / 4 / 5 &  -- \\\midrule
			conv 3 activation & -- & --  \\
			fc 9216$\rightarrow$3072 & 1.3 / 1.3 / 1.5 & 0.9 / 0.9 / 0.43 \\
			fc 3072$\rightarrow$3072 & 0.45 / 0.45 / 0.5 & 0.3 / 0.3 / 0.13  5\\
			fc 3072$\rightarrow$1000 & 0.4 / 0.4 / 0.2 & 0.15 / 0.15 / 0.08 \\\midrule
			Total fc layers &  6.5 & 3.34  \\
			Total all layers &  104.1 \\
			
			\bottomrule[1.25pt]
		\end{tabular}
		\par
		\bigskip
	\end{minipage}
\end{table}

\subsubsection{Analysis of single node 4 GPU parallelism}

With these numbers we can easily calculate the theoretical speedup for AlexNet and with the same reasoning demonstrated in this section we are able to calculate the theoretical speedup for any section.\\
The first thing we can see is that we can completely hide the data parallel communication in the convolutional layers under computation, that is, while we are calculating the gradient for the next convolutional layer we can synchronize the gradients of the previous layer before the computation finishes (sync time $<$ weight update time for any convolutional layer). Only the synchronization of the last layer cannot be overlapped with computation, which means that the synchronization time for this layer is the only cost we pay for the data parallelism scheme in convolutional layers. As we can see, neither 8-bit nor 1-bit compression will yield any speedups in this case and thus we proceed our analysis to the model parallel part.\\

In the model parallel part for AlexNet we follow the scheme introduced in section 2.4, that is we split the convolutional activities in $K$ parts, where $K$ is the number of GPUs, and transfer these sub-batches while we do the forward and backward passes for the previous sub-batches.

More specifically, while we compute the matrix multiplication of the first sub-batch, we can transfer the next sub-batch. Similarly we transfer the fully connected activities (the first true model parallel step) while we calculate the fully connected activities on another sub-batch. We can do this procedure in every layer and thus hide the communication time of the previous layer under the computation time of the current layer. This works for both the forward and backward pass. We can hide the communication of the error activities to the convolutional layers by synchronizing them while we are updating the weights. We can do this except for the first batch of convolutional activities and the last batches of error activities (nothing to overlap with) which incurs a total penalty of $5\times 0.9$ms for 32-bit. Thus for 32-bit we have a total synchronization penalty of:
 \[ (0.9-0.9) + (1.1-0.3) + (1.1-0.15) + (0.4-0.3) + (1.1 - 0.9) + (1.1-0.64) + 5\times 0.9 = 6.81\mbox{ms} \] 
For 8-bit transfers we only have a penalty for some transfers because they mostly overlap fully with matrix multiplication.
 \[ (0.5-0.3) + (0.5-0.15) + 5\times 0.4   = 2.55\mbox{ms} \] 
At the same time we receive a speedup because we are operating on smaller matrices due to model parallelism (see Table 4 for details) which saves us 3.16ms for 32-bit matrix multiplications. From this we can see that model parallelism yields only tiny speedups compared to the model parallel stage. However, we still see some speedups, especially when we use 8-bit approximation.

\subsubsection{Predictions of the single node 4 GPU model}

Putting everything together we now can calculate the theoretical speedup. We use full data parallelism in the convolutional layers and model parallelism of $K=4$ sub-batches through the fully connected layers. In data parallel layers we need the same time as we need for 1 GPU but without the fully connected part which has $K=4$ model parallel passes. On top of this we add all penalties for the model parallel stage and we thus receive the expression:
\[ \mbox{speedup } = \frac{\mbox{Number of GPUs}\times \mbox{total time}}{(\mbox{total time} - \mbox{fc time}) + \mbox{conv penalty} + (\mbox{\#sub-batches}\times\mbox{parallel fc time}) + \mbox{fc penalty} } \]
And this yields in our example:
\[ \mbox{speedup }_{32} = \frac{416.4}{(104.1 - 6.5) + 0.05 + (4\times3.34) + 6.81} \simeq 3.53 \]
\[ \mbox{speedup }_{8} = \frac{416.4}{(104.1 - 6.5) + 0.05 + (4\times3.34) + 2.55} \simeq 3.67 \]
Here the estimated total time of 104.1ms for a full pass is based on convolutional kernels which are about 75\% faster than those used by \citet{krizhevsky2014one}. If we use the total time estimate for the kernels used by \citet{krizhevsky2014one}\footnote{see "convnet2" here https://github.com/soumith/convnet-benchmarks} the model predicts a speedup of:
\[ \mbox{speedup }_{32} = \frac{177\times 4}{(177 - 6.5) + 0.05 + (4\times3.34) + 6.81} \simeq 3.71 \]
\[ \mbox{speedup }_{8} = \frac{177\times 4}{(177 - 6.5) + 0.05 + (4\times3.34) + 2.55} \simeq 3.8 \]

Which is very close, within 1\% to the actual speedup of 3.66 achieved by \citet{krizhevsky2014one}. \\

Our model does not take into account the additional time needed to stack or add the buffers from model parallelism which is typically in the order of 10\% of the time for the respective matrix multiplication. Also we used a different GPU (GTX Titan X). Yet our theoretical model nevertheless predicts the underlying speedup accurately, which should attest to the robustness of our model which we will now extend to a GPU cluster case.

\subsection{A theoretical GPU cluster model for 32 nodes, 96 GPUs}

\subsubsection{Benchmarking components in a 32 node, 96 GPU cluster}

The sub-components of a GPU cluster are similar to a single machine, apart from the fact that we now also have a network with switches between computers. The two bottlenecks in network communication are network bandwidth and network latency. InfiniBand network interfaces usually have very good network bandwidth and network latency and our analysis will be based on a FDR InfiniBand system used in conjunction with MPI software for communication between nodes in the cluster.\\
Benchmarks for InfiniBand systems using MPI are readily available online. Table 5 shows the latency for messages in our model-parallel scheme. To find the InfiniBand latencies we looked at charts that show the latency for a given message size. The message size in this case is the sub-gradient (parameters/32) or the convolutional activities (depends on sub-batch size). From this data we also found that most messages will have a bandwidth of 6GB/s for a FDR interconnect. We use this bandwidth estimate in all our following calculations.

\begin{table}[h]
	\caption{Benchmark data for the model parallelism step in AlexNet for a GPU cluster with 32 nodes and 96 GPUs. }
	\label{sample-table}
	\begin{minipage}
		{\linewidth}
		\centering
		\begin{tabular}{ ccccccc}
			\toprule[1.5pt]
			Layer/activation & Sub-batch size & Forward passes   & \multicolumn{2}{c}{Size (kB)}  &  \multicolumn{2}{c}{InfiniBand latency (ms/msg)}   \\
			& & & 32-bit & 8-bit & 32-bit & 8-bit \\
			\midrule	
			conv3, 384$\rightarrow$256 & --&  1 & 108 & 13.5 & 0.03 & 0.008 \\
			conv 3 activation & 128 & 96  & 144  & 18 &  0.035 & 0.007\\		
			conv 3 activation & 256 & 48   & 288 & 36 &  0.07& 0.01\\	
			conv 3 activation &  512  & 24  &  576 & 72 & 0.1 & 0.02\\		
			conv 3 activation &  1024 & 12   & 1152& 144 & 0.3 & 0.04\\	
			conv 3 activation &  2048 & 6   & 2312& 289 & 0.7 & 0.07\\	
			conv 3 activation &  12288 & 1   & 13872 & 1734 & 5(?) & 0.06\\	
			fc 9216$\rightarrow$3072 & 128  &  1 & 24 & 3 & 0.008 & 0.006\\
			fc 9216$\rightarrow$3072 & 256   & 1 & 48 & 6 & 0.015 & 0.006\\
			
			\bottomrule[1.25pt]
		\end{tabular}
		\par
		\bigskip
	\end{minipage}
\end{table}

\subsubsection{Analysis and speedup prediction of 32 node 96 GPU parallelism}

From Table 5 we can calculate the time needed to transfer the largest convolutional layer during data parallelism. For 2 PCIe transfers of a sub-gradient and subsequent 31 InfiniBand transfers we have two messages of 36 kilobytes at 5GB/s and 31 messages of 108 kilobytes at 6GB/s with 0.03ms latency per message, respectively. We do this messaging scheme twice: Once to distribute the raw gradients, twice so we distribute the accumulated gradient to all nodes. The total time for this gradient synchronization scheme is about 1.9ms. This shows, as in the 4 GPU case, that there is no bottleneck in the data parallelism part of convolutional layers and thus 8-bit or 1-bit quantization will not improve performance in these layers. \\

To look at model parallelism, we first need to benchmark the matrix multiplications for a model parallel pass. We generated new matrix multiplication benchmarks for this case but found that model parallelism in fully connected layers will induce a slowdown because the compute-time for the matrix multiplication (0.1 / 0.08 / 0.08 ) are too small while the latencies alone will destroy performance: $32\times 0.008 = 0.256$ms and $32\times 0.015 = 0.48$ms, for a sub-batch size of 128 or 256, respectively. Different model parallel schemes with speedups are possible, but these are too complex to analyze. Instead we discard model parallelism in these layers and we concentrate on the analysis of the transfer of the convolutional activities to the fully connected layer. However, we still use our sub-batch scheme from section 2.4 in this case.

\begin{table}[h]
	\caption{Benchmark data for synchronization of the convolutional activities in AlexNet for a GPU cluster with 32 nodes and 96 GPUs. }
	\label{sample-table}
	\begin{minipage}
		{\linewidth}
		\centering
		\begin{tabular}{ ccccccccc}
			\toprule[1.5pt]
			Sub-batch size &  Forward time & \multicolumn{4}{c}{Sync time (ms)}   & \multicolumn{2}{c}{Total time (ms) }    \\
			 & &  \multicolumn{2}{c}{Single pass} & \multicolumn{2}{c}{Full pass /w overlap} &  &\\
			 &  & 32-bit & 8-bit &  32-bit & 8-bit & 32-bit & 8-bit \\
			\midrule	
			  128 & 624  & 2.1 & 0.35 &  24.8 & 0.35 & 650 & 624 \\		
			 256 & 312   & 4.2 & 0.55 & 89.9 & 0.55 & 402 & 312\\	
		  512  & 156  &  7.15 & 1.13 & 181 & 1.13 & 337 & 157 \\		
			 1024 & 78   & 17.4 & 2.25 & 499 & 29.5 & 577 & 108 \\		
			 2048 & 39   & 25.5 & 3.26 & 750 & 60.8  & 789 & 100 \\		
			 12288 & 6.5   & 250(?) & 31 & 7750(?) & 921  & 7750(?) & 928 \\				
			\bottomrule[1.25pt]
		\end{tabular}
		\par
		\bigskip
	\end{minipage}
\end{table}

Table 6 shows the timings for the forward passes for different sub-batch sizes. Along with synchronization time this is the main bottleneck in the computation. Table 6 also shows the timings for a single pass of convolutional activities for both 32 and 8 bit and the timing for the full pass (all convolutional activities) where subsequent transfers are hidden under the matrix multiplication of the subsequent layer (1.3ms). The total time is the time of both the forward passes and the synchronization time for the full pass. With these data we receive the predictions for speedups shown in Table 7. Note that AlexNet based on NervanaGPU kernels and convnet2 kernels has a baseline of $96\times 104.1 = 10$s and $96\times 177 = 17$s milliseconds, respectively.

\begin{table}[h]
	\caption{Predicted speedups for AlexNet for a GPU cluster with 32 nodes and 96 GPUs using different convolutional kernels. }
	\label{sample-table}
	\begin{minipage}
		{\linewidth}
		\centering
		\begin{tabular}{ ccccc}
			\toprule[1.5pt]
			Sub-batch size &   \multicolumn{4}{c}{Speedup}    \\
			& \multicolumn{2}{c}{NervanaGPU kernels} & \multicolumn{2}{c}{convnet2 kernels} \\
			& \multicolumn{2}{c}{Baseline 10s} & \multicolumn{2}{c}{Baseline 17s}\\
			&    32-bit & 8-bit			&   32-bit & 8-bit  \\
			\midrule	
			128 & 13.4 & 13.8x		& 20.7x & 21.4x\\
			256 & 20x & 24.4x & 29.7x & 35.2x\\	
			512  & 23x & 39.3x & 33.5 & 51.9x\\		
			1024 & 14.8x & 48.6x & 22.7x & 61x\\				
	   		2056 & 11.3x & 50.6x & 17.7x & 62.8x\\				
	   		12288 & 1.3x & 9.7x & 2.15x & 15.5x\\				
			\bottomrule[1.25pt]
		\end{tabular}
		\par
		\bigskip
	\end{minipage}
\end{table}

\end{document}